# Novel Compliant omnicrawler-wheel transforming module

[1] Akash Singh, [1]Vinay Rodrigues, [1]Enna Sachdeva, [1]Sai Hanisha, [1]Madhava Krishna

*Abstract*— This paper present a novel design of a crawler robot which is capable of transforming its chassis from an Omni crawler mode to a large sized wheel mode using a novel mechanism. The transformation occurs without any additional actuators. Interestingly the robot can transform into a large diameter and small width wheel which enhances its maneuverability like small turning radius and fast/efficient locomotion. This paper contributes on improving the locomotion mode of previously developed hybrid compliant omnicrawler robot CObRaSO. In addition to legged and tracked mechanism CObRaSO can now display large wheel mode which contributes to its locomotion capabilities. Mechanical design of the robot has been explained in a detailed manner in this paper and also the transforming experiment and torque analysis has been shown clearly.

## I. INTRODUCTION

Over the past few decades, there have been numerous studies on the hybrid robots since they can help perform dangerous missions in complex and unpredictable environments, such as rescue, anti-terrorism, removing explosive and so on. For the field service robot, mobility is the one of the main concerns of the researches. There have been many studies about robot mobility for example the study of locomotive mechanism, obstacle traversal, autonomous traveling based on the local and/or global localization, collision avoidance, posture stabilization etc. Hybrid robots capable of performing multiple locomotion modes on variable terrains are very key subject of study for researchers.[1] gives a well-documented literature of the capabilities of wheeled, legged and tracked based ,mobile robots on rugged surface.

Legged robots can navigate on uneven natural terrains as they possess good adaptability by varying the effective length and orientation of the legs [2]. Track robots possess high traction due to large contact surface with the ground and hence have an added advantage over wheeled and legged robots while crawling over holes and rugged terrain since they have the ability to smooth out the path, putting low pressure on terrain [2]. While wheel robots have the edge over other modes in its capability to move fast and navigate efficiently. These potential advantages of each locomotive trait have led to the development of a number of hybrid locomotion based robots [2], [3] and [4]. Identifying the selective capabilities of wheel, tracked and legged robots, we developed a couple of modules based on the hybrid of track and legged robot. In [6], and [8] we presented unique designs of Omni crawler modules, where we introduced active compliance in the Omni crawler [5] module that had been invented earlier. In [7], a modular pipe climbing robot was introduced which utilized the omnidirectional capability of Omni crawler module. Later pipe climbing robot design was improved to introduce active compliance in Omni crawler module in a single plane for constrained space motion. This robot had the capabilities of both a single plane serial manipulator as well as the surface actuation capability of Omni crawler. Further in [8] we developed CObRaSO (Compliant Omni-Direction Bendable Hybrid Rigid and Soft OmniCrawler Module) which is a compliant omni-bendable omnicrawler module, capable of performing all three locomotion modes. CObRaSO had its advantage over previous hybrid designs that the same module could behave as a hyper-redundant manipulator (behaving as legged robot as in quadruped, snake like robot) as well as a compliant Omni crawling robot with high traction surface. Also the thin circular cross-section of the robot provided additional omnidirectional locomotion functionality. CObRaSO presented its unique motion configurations individually and it showed its scalability to a large number of diversified structured robots.

[1]All authors are with Robotics Research Center, IIIT-Hyderabad, akashvnit2016@gmail.com

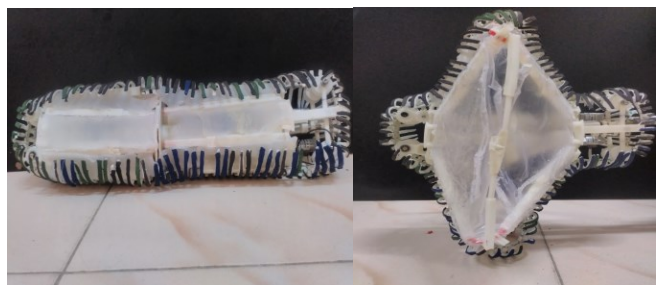

Fig. I: Transformable compliant Omni crawler robot

CObRaSO module to quite an extent resembles a skin actuated snake robot having the capabilities of high traction motion using crawler and also having a 4 DOF internal compliance. Hyper-redundant skin actuated mechanisms consisting of a series of joints which move via internal shape changes like a snake [9] as well as a high traction track robot. However such skin actuated snake robots (including CObRaSO) modules are not capable of performing highly maneuverable locomotion modes that can be exhibited by wheeled robots. Maneuverability is a very important aspect in the case of compliant snake like robots. They are given control inputs to perform locomotion gaits on even surface in order to turn in place/differential turn or moving sideways. Moving forward, steering and lateral motion challenges were addressed by the compliant Omni crawler modular robots but when it comes to turning with zero radius, compliant crawler too faces challenges as it requires large space to turn. Skidding on the surface also occurs in differential turning since the area in contact with ground surface is quite large. The radius of turning is typically large for modular robots made using CObRaSO such as snake robot, quadruped (in planar configuration) as shown in Fig. 1. Hence there is a need of introducing high maneuverability of wheeled robots to the compliant Omni crawler such that it can revolve on a place in thin space and without much skidding. Generally, transformable track robots have application in various types of environments. For example, CALEB-2 [9] a compliant tracked robot is capable of handling uneven terrains, ROBHAZ-DT [10] is able negotiate rough terrain negotiable mobile platform with passively adaptive double-tracks, and Single-Tracked [11], Azimut[12] are some typical transformable track robots robots which can exhibit tracked as well as wheel modes. Most of these robots have bulky tracks and even after transformation at any stage of their operation the length and size of the robot remains large. Hence the maneuverability remains less as they require large turning radius. Also in these robots the tracks run over the sprocket in all motion configurations [14], hence locomotion efficiency remains less. In this paper we present a transformable CObRaSO module which can operate as both an Omni-compliant crawler robot as well as a large size wheeled robot. The transformation between the two modes occurs without any addition of extra actuator and attachments. Interestingly the robot has an additional advantage of having a variable wheel size which can let it slip though constrained heights and also help it climb over obstacles. This paper focusses on the novel design of a crawler wheel transformable robot which displays the ability of both a long high traction tank belted robots as well as highly maneuverable wheeled robots.

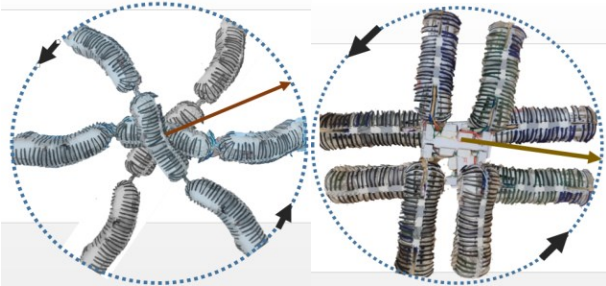

Fig.1: Shows the turning radius of modular snake robot and planar quadrupeds made using CObRaSO.

## II. ROBOT CONCEPT

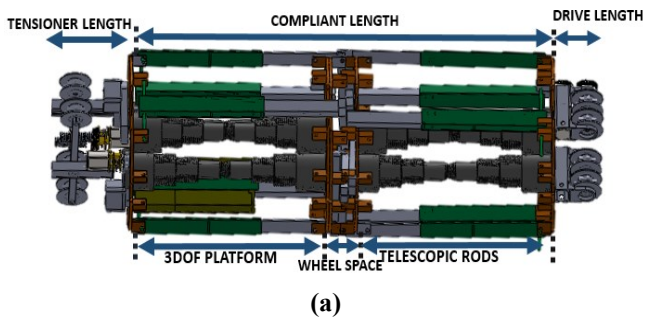

(a)

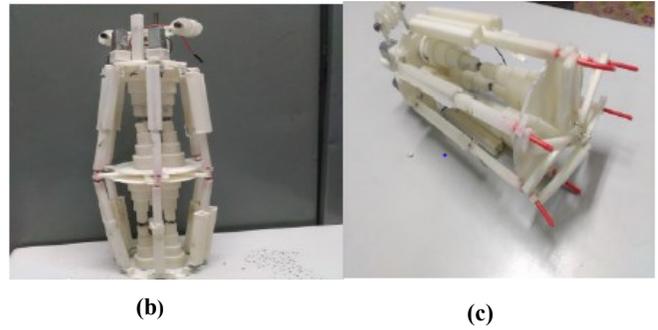

(b)          (c)

Fig. 2: (a) The CAD model of internal cage displaying different section lengths of the robot, (b) shows the actual prototype of the module placed vertically, (c) shows the cross-section of the robot at the wheel space location as shown in Fig 2.a.

The robot consists of a novel shape changing mechanism which is inspired by transforming the chassis of the robot into a wheel. This means that the chassis of the robot is used for dual application in which in one case it acts as a surface over which the chain-lug combination of the crawler module runs over the surface of the robot maintaining strict contact with chassis overall. The module in this mode is a thin circular cross-sectioned but long structure which provides high traction with the ground, behaving like a surface actuated snake robot. The robot manufactured using this module is capable of moving in tight space and the omnidirectional nature of the module design helps moving laterally and also using omnidirectional compliance enables it to perform high traction motion even in lateral direction. Though such robot structures have massive advantages but compliant robots face tough challenges in moving over obstacle due to their thin cross-section. Also challenges like large turning radius, restrict its maneuverability in tight spaces. Hence to overcome these limitations the robot transforms into a big diameter and thin sized wheel. This transformation helps decrease the length of the module by more than fifty percent of its ideal crawling length. This wheel mode enables the robot to turn in a very confined space, enhancing the capability of any kind of robots like snake robot, quadruped or even biped developed using this module. Reduction in the length of a compliant robot like CObRaSO without involving additional actuators and keeping the compliant quality of the robot intact is the key contribution of this paper. The joints used in previous design of compliant tracked robots consist of revolute actuated joints. The previous design of CObRaSO consisted of a series of actuated joints interconnected with each other using connecting plates. This design structure enabled the robot to have compliance in both the directions. But total length of the median line passing through the center of the connecting plates of the robot remained constant. In order to change the length to vary length of the chassis of the robot as well as maintaining the compliant nature of the robot led us to come up with novel design structures like telescopic 3 DOF bending platform using linear actuator which help changing chassis length using individual power screw mechanisms. The need to reduce the length of the robot led us to innovate upon the existing mechanisms and also introduce new mechanisms including triggering mechanism, which mainly switches the robot modes, spoke wheel connection mechanism, which helps maintain a continuous outer surface of the wheel in the wheel mode, which otherwise remains intact in a capsule in the compliant-track mode. The transformation of the robot externally is similar to that of the transforming origami robot in visually. But origami robots are soft in nature and unsuitable to handle high payloads and rough locomotion modes. For the transformation between two modes to occur using limited actuators and keeping the robot body hard(and not a soft body like that of an origami), there arises a need of transforming of long robot parts to short robot parts and vice versa lengths in small. All these mechanisms work synchronously to achieve the novel transformation. While it's very interesting to acknowledge that all these mechanisms involve the usage of telescopic mechanism in some way or the other formats. The 3 DOF parallel bending platform uses the telescopic screw mechanism, while other two mechanisms use straight telescopic rods and curved telescopic rods respectively. This design helps transforming the robot with the same set of actuators that are involved in providing compliance of the crawler robot. Apart from this, this paper also discusses the few design improvements that have been done over previous robot design considering compliance/transforming part of the robot to be as large as possible. The next few sections of the paper focus on the detailed design of the robot, explaining the different design parameters in detail. Section IV focusses on the output torque calculation of the motors involved in providing compliance and transformation. The results for design parameters and the attained diameter of the wheel after transformation are given.

## III. ROBOT DESIGN

The Design of the robot consists of a central telescopic 3 DOF platform mechanism which is cascaded with an identical platform to provide higher degrees of bending along with a chain tensioner mechanism to maintain the constant tension in the chain, when the module is bent in any direction. The chassis of the robot consists of 6 pairs telescopic rods arranged at 60 degrees with respect to each other which surround the 3 DOF compliant mechanisms. The rods in each pairs of rods are connected via compliant thick wire connectors. The driving mechanism consists of a driving motor connected to driving shaft via gears.Fig.2 shows the exploded CAD model version of the module design. Wheel connector mechanism connects all the rods using curved telescopic rods \to form the entire shape of the wheel. The different parameters of the mechanisms are explained as follows.

### A. Cascaded Telescopic 3 DOF Platform

The robot compliance mechanism design has to be such that the expansion reduction size is maximized as well as the total length of the robot is large enough to maintain high traction on slippery surface and move over uneven terrain like stairs and pot holes. In order to keep the higher degree of freedom

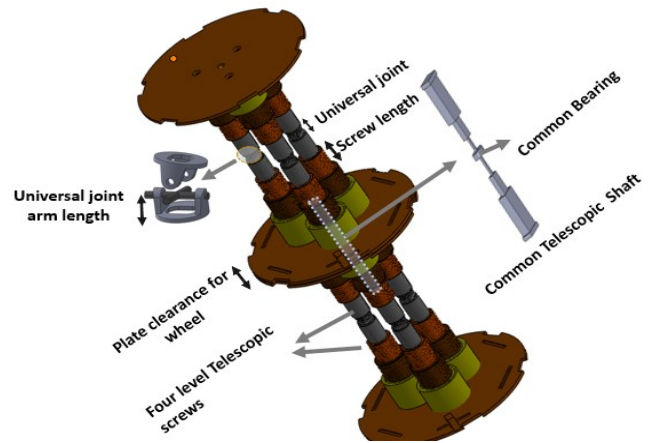

Fig.3: (a) Shows the detailed CAD model design of the cascaded telescopic 3 DOF manipulator, depicting the internal common shaft and universal joint.

The Omni-directional bending is realized with 3 'Telescopic screw mechanism' based linear actuators, aligned at 120_ apart from each other on a circular rotating plate. Previously, the Double-Screw-Drive (DSD) mechanism has been proposed by Ishii, Chiharu, et al [13], where the Omni-directional bending is achieved with 2 rotating linkages with left & right-handed screws and universal joints. The desired bending angle proportionally increases the length of the screw, thereby increasing the size of the coupler. In order to achieve sharp bending while keeping the size of coupler within constrained space, a 'Telescopic screw drive mechanism' is introduced. It interconnects multiple screws, similar to rods in a telescopic antenna. It consists of a master screw connected directly to the actuator's rotor and the outer threads of successive screw are fastened to inner threads of the preceding screw. When the master screw reaches its extreme end during actuation, the stopper at its end transfers the rotational motion to the successive screws in a sequential manner. The arrangement of the screws, universal joints and the circular rotational links in one sub-module is shown in Fig 5. The robot consists of a cascade of such sub-modules and the power imposed on each of the telescopic screw assembly of a sub-module is transferred to the corresponding assembly of the cascaded sub-modules. Therefore, the number of sub-modules that can be cascaded is limited by the maximum power rating of the actuators. The cascaded platform is actuated by 3 actuators and rotational motion from one platform to the other is transferred via common shafts, which internally couples the left handed master screw of the first platform with the corresponding master screws of the second platform. The differential Rotations of the 3 telescopic screws result in the omnidirectional bending of the module. The telescopic screw mechanism we have developed is a level 4 mechanism in which the each level is of length 2 cm. The stopper size is 2mm in height. At its full length the telescopic set can increase to a height of 80mm. The maximum extension and reduction of each of the 3 DOF manipulator platforms and cascaded 3 DOF platforms is given in equations (1) and (2). The total reduction in the length of the robot is also derived.

| Symbols | Value |
|---|---|
| $N$ | Number of Telescopic levels of screw |
| $S_L$ | Length of screw including stopper |
| $J_H$ | Total height of Universal joint. |
| $D_L$ | Chain driving motor assembly length |
| $T_L$ | Chain tensioner assembly average length |
| $U_{AH}$ | Universal Joint arm height |
| $C_l$ | Clearance length between the adjacent plates |
| $L_E$ | Original length of module |
| $L_R$ | Reduced length of module |
| $Ns$ | Number of Telescopic levels of common shaft |
| $T_w$ | Screw Thread width |
| $Tc$ | Inter screw thread clearance |
| $D_N$ | Outer diameter of Nth level telescopic screw |
| $Sw$ | Width of stopper between two adjacent screws in telescopic arrangement |

The level of the telescopic screw was decided by deciding the amount of length reduction that needs to be met. The total reduction in the length of the robot considered to be more than half the elongated length for the testing of the design.

$$L_E = ((N*S_L)*2)*2 + (J_H)*2 + (C_l) + D_L + T_L \qquad (1)$$

When the robot is in maximum reduction mode, the reduced length of the robot, $L_R$

$$L_R = ((S_L)*2)*2 + (J_H)*2 + (C_l) + D_L + T_L \qquad (2)$$

$$J_H = 2*(U_{AH}) \qquad (3)$$

For reduction of more than half to happen,

$$L_R / L_E < 1/2 \text{ or } L_R / L_E = 1/2. \qquad (4)$$

From Eq. (4) and (2),
Reduction of the length is aimed to be more than half of the total robot length. By considering $L_R / L_E = 1/2$, and substituting in eq. (2), the value of the minimum screw length for each level is derived, considering screw lengths as equal.
Due to design constraints, like the minimum size of high torque motors, and 3d printing precision limitations, the driving and tensioner length of the robots are kept fixed to as minimum as possible. Putting the constraints of driving length and tensioner mechanism. We get the level of the telescopic screws as 4.
$$N=4; \qquad (5)$$

Each of the 3 DOF platform are interconnected using a telescopic internal shaft. This ensures that both the platforms rotate simultaneously. The number of level of telescopic shaft is determined by the number of the levels of screws.
Hence number of internal shaft telescopic levels, Ns

$$Ns=N-1=3; \qquad (6)$$

The major design consideration to be taken care of for this robot is to make the cross section of small as possible. The internal shafts are made using Electric Drill method, which allows a minimum of 1.5*1.5 square millimetres of holes to be cut. The internal screw will accommodate all the telescopic shafts when they are reduced to the minimum length. Apart from this considering minimum widths of screw threads and stopper mechanism between each telescopic levels, made us arrive at the diameters of the screws of all telescopic screws as,

$$D_N = D_{N-1} + Tw + Tc + Sw. \qquad (7)$$

$$D_1 = 1.5 + 0.5*(Tw) + 0.3(Tc) = 2.3 \text{mm}. \qquad (8)$$

$$Tc = 0.5 \text{mm}. \qquad (9)$$

The above equation is helpful in deciding the minimum diameter of the outermost screw effecting the arrangements of the screws, and eventually the minimum diameter of the chassis. Telescopic screws aligned 120 degrees between two circular rotational plates and are paced as close as possible to induce maximum bending in minimal rotations of the screws in the compliance mode.

### B. Design of the Telescopic Chassis Rods

Designing the chassis of the robot which maintains a solid outer surface during compliance in any direction, during robot transformation as well as in the wheel mode is also one of the key contributions of this paper. This is to be noted that the chassis design is such that the transformation from track to wheel mode consumes minimum number of actuators. Some important points taken into consideration while designing the chassis were:

1) The chassis should be as rigid a surface as possible in any configuration. This ensures that the robot remains robust and makes it capable of overcoming uneven terrains.
2) The chassis should accommodate the path of the chains in the tracked mode such that the lug-chain combination always remains in high tension and always be compliant with the shape of the robot.
3) The chassis should also be compliant in order to maintain the functioning of the tracks even in bent configurations.
4) During transformation between the tracks to wheel mode the chassis should deform bulge out radially equally such that the wheel has perfect circular surface.

In order to meet the above considerations the chassis was designed as a combination of hard and soft structures such that the surface remains continuous even during reconfiguration. Continuous surface ensures that the lugs and chain assembly always have a surface to slide upon. Six pairs of 2 level telescopic rods are used as the core structure of the robot chassis. The rods are connected to a common circular disc plate, such that the arrangement creates a cage like structure as shown in fig. 4. The telescopic rods in each pair are connected together via a thick flexible wire. This joint is spherical in nature and has a very large bending range. The joint is located very near to the middle plate of the robot. The length of the telescopic segments of the rod is decided by the maximum bending angle of the platform. The central joint of each pair is fixed to the middle plate in the compliant track mode.

In order to maintain a continuous surface the telescopic rods are interconnected via flexible silicon sheet of the very smooth material Ecoflex 00-20. The sheet design is derived from the design in [3] where two adjacent rods are interconnected to the silicon sheet, the silicon section depending upon the radial position on chassis expands or contracts when the module bends in any direction or perform the transformation motion. The telescopic design of the robot depends upon two parts as shown in Fig.4.

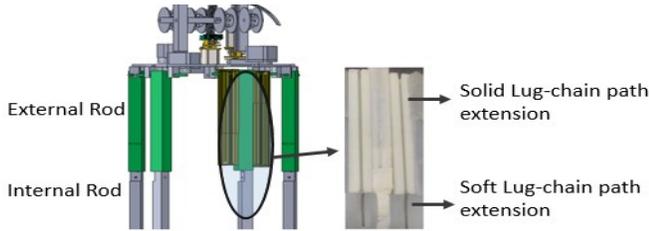

Fig.4: Solid and soft lug chain extension path and the telescopic rods which form the circular cage (chassis) of the robot

In order to satisfy the second condition for the chassis design, two diametrically opposite rods of the six telescopic rods of the chassis are modified in design and are connected to the solid chain track extensions as shown in fig. 4 (Solid Lug-chain extension). The length of this extension is limited to the length of the outermost telescopic rod .This chain track extension is connected further longitudinally to silicon track of n identical cross section to that of the rigid track. Therefore, in the proposed design the shape of the robot is controlled by rigid actuators and the soft silicone rubber body passively deforms according to the design constraints of the rigid body. The cascaded arrangement of silicone rubber and rotating plates provides a continuous pathway for lug-chain assembly in straight as well as bent configuration, as shown in Fig. 4. The silicon chain track are at one terminal connected to spherical joint connecting the rods placed at the middle plate of the robot chassis.

### C. Calculating the Diameter of the robot

The diameter of the robot is indirectly dependent upon the maximum bending compliance of the robot. Each telescopic 3 DOF (Degrees of freedom) motion can be represented as the motion of two cascaded triangular plates as shown in fig 5. The bending of the robot is uniform throughout its 3 DOF manipulator length. And hence as shown in the fig. 5(Right), we arrive at the equation as follows,

$$\Theta_{1/1}= \Theta_{2/2}= \Theta_{3/3}= \Theta_{4/4} \quad (10)$$

$$\Theta_4 = 4*\Theta_1. \quad (11)$$

In order to comply through sharp turns and to give an output angle identical to that of an active revolute joint in previous snake like compliant robots, we consider the total bent angle to be 90 degrees. Hence,

$$\Theta_4 = \pi/4; \; \Theta_1 = \pi/16. \quad (12)$$

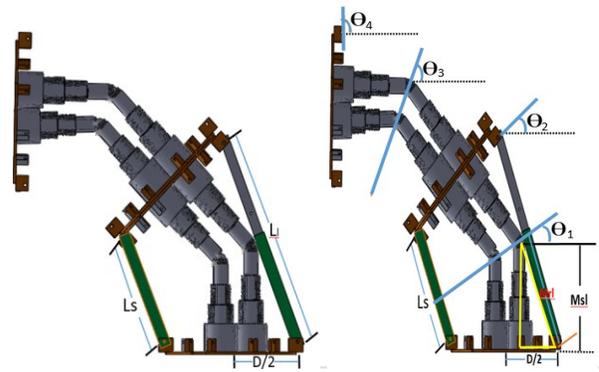

Fig. 5: Right: CAD model depicting different angular planes formed due to cascaded bending of dual 3 DOF platforms.

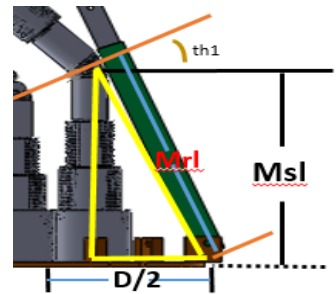

| Symbols | Value |
|---|---|
| $V_s$ | Component of the distance screw from the centre on the line joining the centres of the any other screw and the centre point of the robot. |
| $M_{sl}$ | Maximum telescopic screw expansion length |
| $D$ | Diameter of the base plate of Telescopic 3DOF manipulator platform |
| $B_l$ | Base length of the angled triangle highlighted in bright. right |
| $L$ | Distance between the centres of adjacent telescopic screw |
| $M_{rl}$ | Half of the maximum telescopic rod expansion length |
| $I_{rl}$ | Internal rod length of the Telescopic rod |
| $E_{rl}$ | External rod length of the Telescopic rod |

In order to calculate the diameter, D of the robot we need to calculate the base length of the right angled triangle highlighted in bright.

$$D/2 = B_l + V_s \quad (13)$$

$$V_s = L*\cos(\pi/3) \quad (14)$$

$$B_l = M_{sl}*\sin(\Theta_1) \quad (15)$$

In this design, the distance of the screws from the centre is given by constraints mentioned above, here L=24mm. $M_{sl}$ value is calculated by using eq. (1), as substituting by 80mm. This provides the value for the D according to eq. (10) as 94mm.

$$M_{rl} = (W_d + d/2)*\sin(\Theta_1) + M_{sl}*\sin(\Theta_1) \quad (16)$$

Where, 'd' is the diameter of the universal joint.
The maximum length of the individual telescopic rod, $T_{lmax}$ is given by the equation:

$$T_{lmax} = 2*M_{rl} \quad (17)$$

Minimum length, $Tl_{min}$ of the telescopic rod,

$$Tl_{min} = Tl_{max} - 2*(D/2)*\sin(\Theta_1) \quad (18)$$

Hence the length of the telescopic screw segments is

$$E_{rl} = T_{lmin} \quad (19)$$

$$I_{rl} = T_{lmax} - T_{lmin}. \quad (20)$$

Substituting 'D' derived from previous equations, and putting $W_d$=5mm and d=2.5mm,

### D. Triggering Mechanism

The telescopic rods which form the chassis behave differently in different modes. In the compliant track mode the rods function as telescopes performing expansion and reduction of their length. But during transformation motion and during wheel mode they behave as rigid and fixed length rods because the in order to perform as wheels the rods behave as spokes, and spokes need to be rigid to support the shape wheel outer diameter during rolling motion.

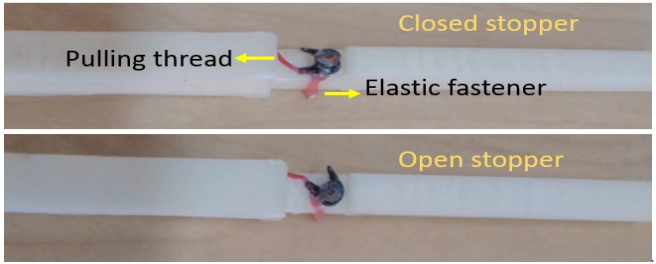

Fig. 6: Prototype showing different components and positions of stopper

The triggering mechanism consists of a circular rotation R shaped slot. The R is hinged using a pin though its central hole. The R shaped stopper is passively hinged to its original zero position via an elastic attachment connecting one leg of R with the inner telescopic rod. In the normal functioning mode of the tracked robot the R shaped slots are at their zero position resting there with the help of elastic attachments. Hence the rod is able to behave normally like a telescope. But during transformation of the module to wheel, these rods become rigid and act as spoke. This occurs by actuating these R shaped slots together by pulling them using a thread whose one end is connected to the leg of R and one the other end are attached to the pulley of triggering actuator. The triggering mechanism is attached to each of the telescopic rods and the thread attachments are all commonly connected to the pulley of the triggering actuator.

When the triggering actuator is powered all the threads attached to the R-as shown in Fig.6. The threads are arranged and attached to all the R stoppers in such a way that at the normal extended length of the robot, the pull threads are taut and triggering actuator is effective as soon as the robot attains its original straight length. Hence eve Ө n after even after many transformation processes. Hence when the robot goes through transformation from original track length to wheel, the R-stoppers are triggered, they stop the external telescopic rod to slide over the inner rod from their original position, hence the telescopic rod behaves as a rigid body, during transformation the R stoppers need not be pulled since the reactions force on the wire joint during transformation pushes the rods towards each other keeping the R stopper always at one extreme of its rotational degree range as shown in Fig .6

### E. Wheel Transforming Mechanism

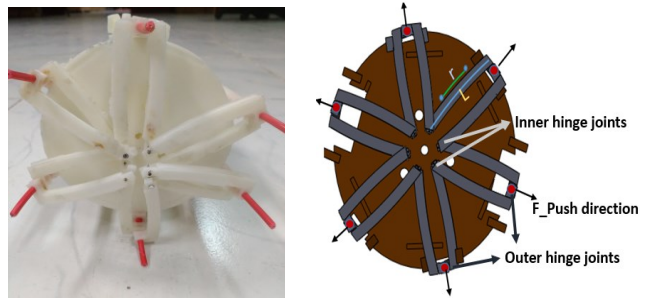

Fig. 7: Left: Prototype of wheel transforming mechanism
Right: CAD design showing the flower type structure and direction of forces on hinges due to wire joints.

Once the transformation has taken place from compliant-track mode to wheel mode, the chassis rods bulge out to form a symmetric spokes arrangement structure. This arrangement is not close to wheel and its outer perimeter is discrete .This structure is unsuitable for smooth motion on flat surface. So to transform into a fully functional continuous wheel, wheel transforming mechanism is used.

This mechanism consists a set of two level telescopic curve rods symmetrically arranged radially as shown in the figure .7 (Right). Any telescopic rod is attached to two adjacent rod, where the outermost rod is connected most rod is connected to an outermost rod of an adjacent telescopic rod through a hinge joint at the perimeter of the central circular plates of the module, while the inner most screw is connected to an innermost screw of other adjacent telescopic rod via a passive revolute joint. The axis of the passive revolute joint at the perimeter of circular plates between outermost screws is coincided by the thick wire of the common wire joint between opposite telescopic rods of the chassis. Hence during transformation the wire joint pushes the outer joints located at the perimeter as shown in Fig.7 (Right) to bulge out and subsequently pulling out the internal hinge joint located near the center of the all the connected pairs.Fig.9 shows the functioning of the mechanism, rod 1 and rod2 are connected to the outer passive joints pull out the inner joint.

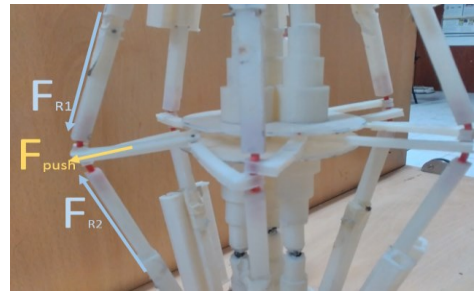

Fig 8: Shows the forces due to rods R1 and R2 cause a resultant outward force on the outer passive hinge joint.

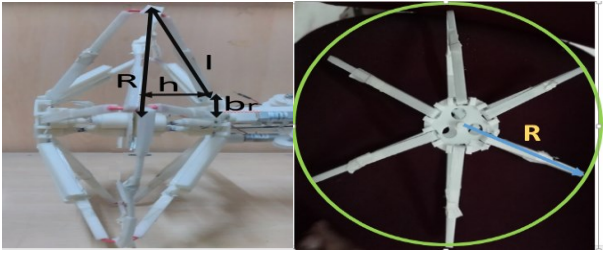

Fig .9: Left: Side view of fully expanded spoke structure
Right: Top view of spoke structure with wheel transform mechanism.

During transformation when the two terminal plates of the module come close to each other and the rods bulge out pulling the inner most joints located at the center to form a circular cage. As shown in fig the radius of curvature marked is calculated by

$$R = ((l^2-h^2)^{1/2} + b_r) \quad (20)$$

Since the flower like structure in the fig. 7 unfolds/elongates to its full telescopic size to become the large circle of radius R, hence

$$r=R \quad (21)$$

Another parameter which is to be taken care of is the number of telescopic levels designed for the curved rod with radius of curvature r and length L such that after transformation, the curved rods extend to become segment of the large circle coinciding the wheel. Length of segment of circle between two adjacent rod wire joints, $S_{l2}$

$$S_l = 2*\pi*R/6. \quad (22)$$

$$S_l = N_{cr} * (L-h_o). \quad (23)$$

$$N_{cr} = S_l / (L-h_0) \quad (24)$$

In the above equation $h_o$ is the rod length utilized for hinge joints attaching the adjacent telescopic inner rods near the center.

### F. Chain Tensioner and Driving mechanism

To keep the chain intact with the bent chassis of the module and maintain optimal tension in both the chains while crawling in bent position, a mechanism to adjust the tension has been realized using linear screw actuators. The actuator adjusts the position of one of the sprockets corresponding to a chain, with a slider. As the linear screw actuator rotates, the sprocket slides along the slider and the tension of the corresponding chain is adjusted. The arrangement of linear, screw actuator, sprocket and the slider are shown in Fig. 8.

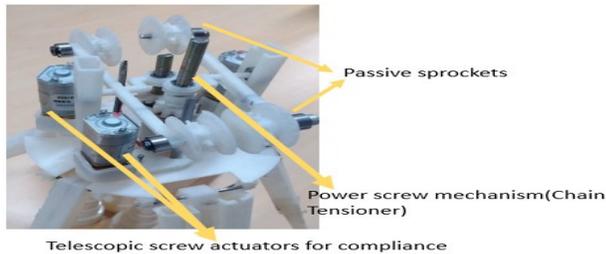

Fig .10: Top: View of fully Chain tensioner section of module;
Bottom left: Driving mechanism with chain; right: Chain tensioner

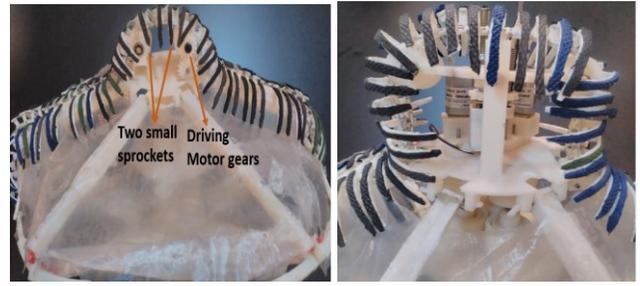

Fig .11: left: side view of transformed wheel module; Right: Top view of transformed wheel module

The driving mechanism is similar to that of CObRaSO module design .In order to keep the relative length of the driving mechanism and chain tensioner mechanism as small as possible, the single each large sized driving sprockets have been replaced with a pair of small sprockets placed at a distance from each other.

### G. Motor Torque calculation

A conclusion section is not required. Although a conclusion may review the main points of the paper, do not replicate the abstract as the conclusion. A conclusion might elaborate on the importance of the work or suggest applications and extensions.

## IV. EXPERIMENTS

The testing of the mechanism is done by powering all three actuators (polulu 15kg-cm micro motors) of the telescopic cascaded 3DOF platform. During transformation the silicon connecting the chassis rods starts expanding in area since the distance between terminal of adjacent rods increases.

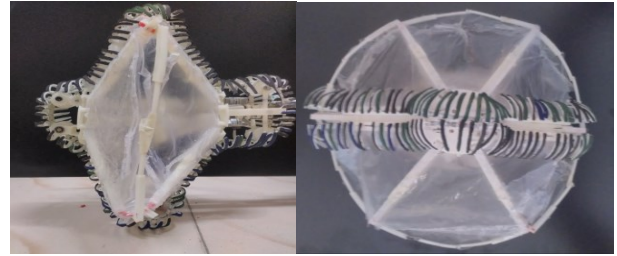

Fig.11. shows the reduced length module to almost its half.

Parameters like maximum permissible torque, gear transmission ratio, operating voltage, temperature, design limitations such as size and cost., determine our choice of motors. Maximum torque of the motor required during the functioning of the robot should be well within the stall torque capacity of the motor. The estimated torque during its transformation from crawler mode to wheeled mode is obtained by simulating the model in ADAMS MSC.

In order to validate the torque and study the behaviour of the transformable robot, simulation is visualised in ADAMS and SolidWorks. The internal telescopic shaft powered by the motors, drives the robot. During the initial stage it is in crawler mode as shown in fig. On powering the screws, a decrease in robot length and an increase in diameter can be observed as shown in figure. With a compression length of we arrive at a wheeled robot as shown in fig. 12(a) & 12(b)

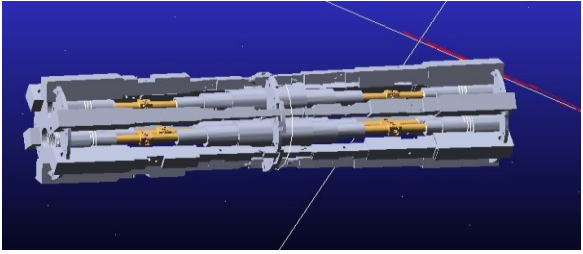

Fig. 12(a): - Crawler mode

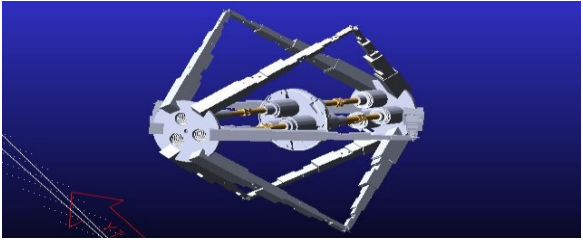

Fig 12(b): Wheeled mode

The chassis of the robot is covered by silicon in order to provide support to the structure and the track. With the change in surface area, silicon being a soft material exerts an internal compressive force on the chassis opposing the transformation. Net moment and force due to silicon on the center of mass of the body is zero, but these results in an internal change in shape. In order to attain the value of the force exerted by silicon, we have designed an experiment imitating the same change in surface area of the silicon. Silicon is wrapped along the chassis with no internal components, and a similar change in diameter and length is obtained by a horizontal force on the base. The horizontal force enables the system to be in a quasi static state, and this is equivalent to the net opposing force by the silicon towards the center. It is measured by using a balance on one side of the base and a rigid support on the other end. Experimentally determined horizontal force at various stages of transformation is given in the Table 3:

The resultant maximum force of 3.4N, applied by the silicon is obtained during the initial steps of transformation. It is the time where there is a maximum change in area. During the entire transformation, maximum force due to silicon is 3.4N. Incorporating this value in ADAMS simulation, gives us the upper range of torque required by the three motors. Graph given below gives us the trends in motor torque in Joint 1

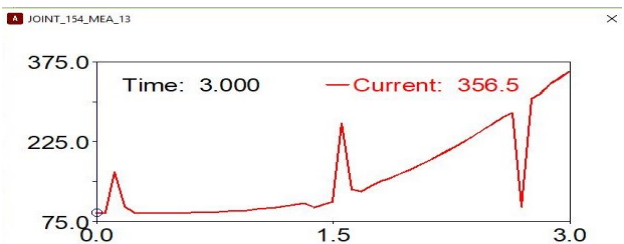

Fig 13: Torque of Motor Joint 1(y-axis (Torque-N mm), x-time)

Selected Motor Torque should always be greater than the maximum torque obtained from the simulation. Torque should be greater than the maximum of the highest torque for the motors. I.e. T> 0.5N.m

| Symbols | Value |
|---|---|
| R | Maximum Diameter of the wheel = 400mm |
| L | Normal length of chassis rod = 140 mm |
| D | Diameter of the base plate of Telescopic 3DOF manipulator platform = 80 mm |
| L | Distance between the centres of adjacent telescopic screw |
| Mrl | Half of the maximum telescopic rod expansion length= 80 mm |
| Irl | Internal rod length of the Telescopic rod = 85 mm |
| Erl | External rod length of the Telescopic rod = 75mm |
| $L_E$ | Original length of module =340 mm |
| $L_R$ | Reduced length of module = 165 mm |

| Change in Length (cm) | Force (N) |
|---|---|
| 1 | 3.4 |
| 2 | 3.2 |
| 3 | 2.5 |
| 4 | 2.1 |
| 5 | 1.5 |
| 6 | 1.0 |
| 7 | 0.6 |
| 8 | 0.1 |

Table. 3

## V. CONCLUSION

In this paper we introduced a novel design of a transforming crawler module. The module is capable of changing its length size over a wide range. The compliant behavior of the omnicrawler module is now supplemented with high mobility of wheels. Hence now the robot is capable of turning in tight spaces, showing high maneuverability using wheel mode as well as can be capable of overcoming obstacles in the path easily. Hence now the CObRaSO module is capable of performing legged, tracked and wheeled motion. The upgradation of the locomotion capabilities of the soft/hard CobRaSO module has also been tested and stated that the efficiency of the new big wheel mode is more than even the wheel mode exhibited by the previous hybrid CObRaSO module.